\documentclass[letterpaper]{article} 
\usepackage{aaai2027}  
\usepackage[hyphens]{url}  
\usepackage{graphicx} 
\usepackage{natbib}  
\usepackage{caption} 
\usepackage{algorithm}
\usepackage{algorithmic}
\usepackage{subcaption}
\usepackage[table]{xcolor}
\usepackage{makecell}
\usepackage{newfloat}
\usepackage{amsmath}
\usepackage{listings}
\DeclareCaptionStyle{ruled}{labelfont=normalfont,labelsep=colon,strut=off} 
\floatstyle{ruled}
\newfloat{listing}{tb}{lst}{}
\floatname{listing}{Listing}

\usepackage{booktabs}
\usepackage{multirow}
\nocopyright

\title{DocTrace: Towards Traceable Long Document VQA via Hierarchical Evidence Graph Reasoning}
\author{
    Le Xiang\textsuperscript{\rm 1}\equalcontrib,
    Zhicheng Guan\textsuperscript{\rm 2}\equalcontrib,
    Hong Chen\textsuperscript{\rm 1}\corresponding,
    Xiaocong Lin\textsuperscript{\rm 1},
    Zhenghua Lei\textsuperscript{\rm 1},
    Teng Hu\textsuperscript{\rm 1},
    Bolei He\textsuperscript{\rm 1},
    Long Zeng\textsuperscript{\rm 2}\corresponding
}
\affiliations{
    \textsuperscript{\rm 1}Baidu Basic Model Research and Development Department, Baidu Inc.\\
    \textsuperscript{\rm 2}Tsinghua Shenzhen International Graduate School, Tsinghua University\\

    xiangle@baidu.com, gzc24@mails.tsinghua.edu.cn, chenhong13@baidu.com, zenglong@sz.tsinghua.edu.cn
}

\begin{document}

\maketitle

\begin{abstract}

Long Document Visual Question Answering (LongDocVQA) requires Multimodal Large Language Models (MLLMs) to locate, integrate, and reason over heterogeneous document elements distributed across multiple pages. Existing approaches, including end-to-end MLLMs, retrieval-augmented generation (RAG) pipelines, and document agents, often lack explicit mechanisms to represent and verify how grounded evidence is progressively composed during reasoning, limiting both answer accuracy and traceability. In this paper, we cast LongDocVQA as an explicit evidence graph reasoning problem rather than implicit answer prediction. To this end, we propose \textbf{DocTrace}, a hierarchical framework that progressively performs evidence localization, structured document parsing, and evidence graph reasoning to enable explicit evidence provenance. To effectively learn these capabilities, we develop a two-stage training framework: joint Supervised Fine-Tuning (SFT) first initializes evidence localization and graph reasoning abilities, followed by task-specific Group Relative Policy Optimization (GRPO) with dedicated rewards to further optimize these capabilities. Extensive experiments on MMLongBench-Doc, LongDocURL, and SlideVQA demonstrate that \textbf{DocTrace} consistently outperforms both existing open-source baselines and proprietary MLLMs. Compared with the Qwen3-VL-8B-Instruct backbone, DocTrace achieves absolute improvements of {\bf 14.4}, {\bf 11.3}, and {\bf 11.7} points on the three benchmarks, respectively. Beyond competitive performance, \textbf{DocTrace} constructs traceable evidence graphs with explicit node-level provenance, enabling transparent and verifiable reasoning for long document understanding.

\end{abstract}


\section{Introduction}

\begin{figure}[t]
\centering
\includegraphics[width=0.9\columnwidth]{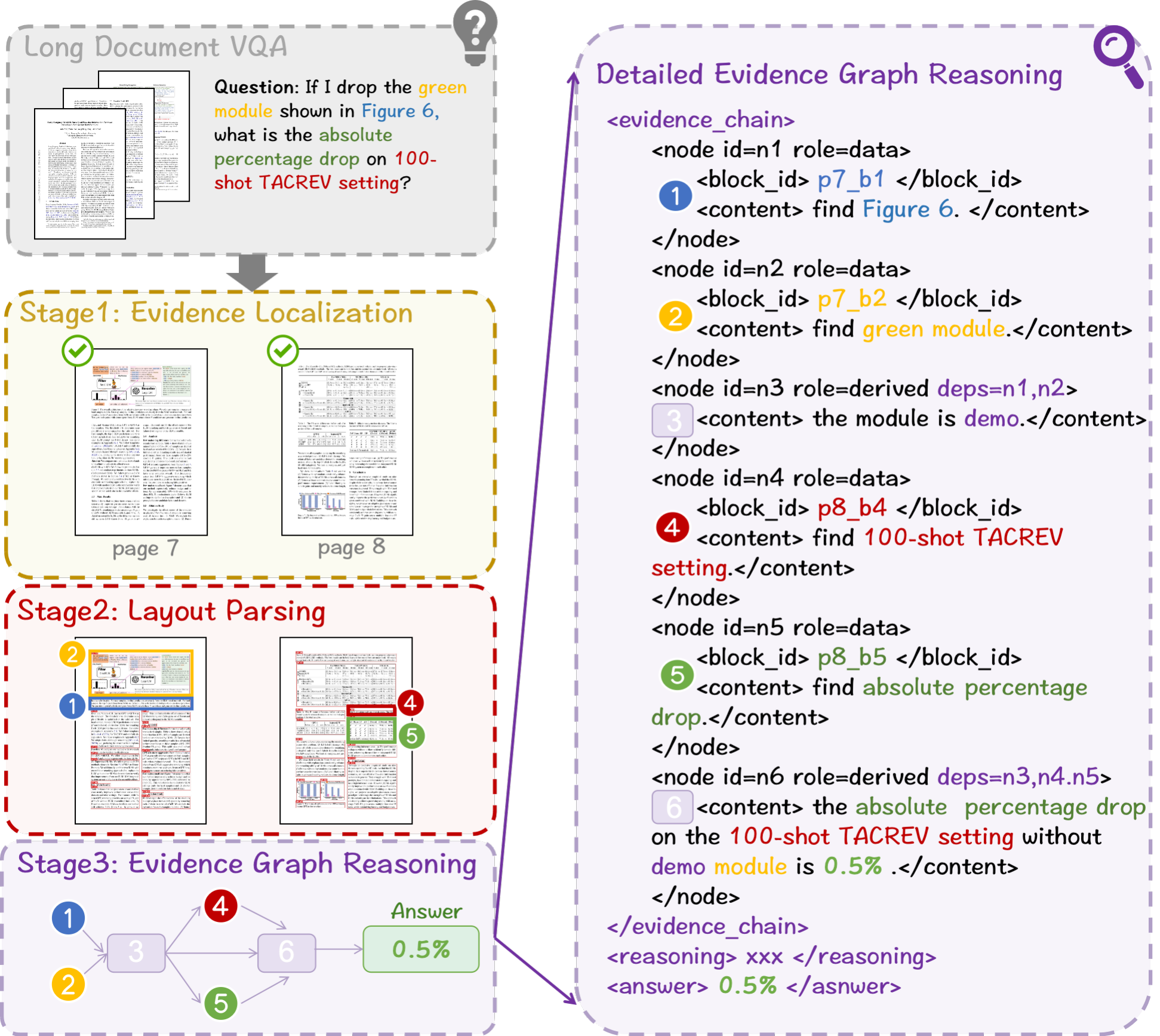}
\caption{Overview of DocTrace inference and its traceable evidence graph. When solving a LongDocVQA problem, the final answer remains fully traceable back to its source nodes and document pages.}
\label{fig1}
\end{figure}

Long Document Visual Question Answering (LongDocVQA) requires Multimodal Large Language Models (MLLMs) to locate, integrate, and reason over evidence dispersed throughout long, visually rich documents containing heterogeneous elements such as texts, tables, charts, and figures~\cite{tito2023hierarchical,ke2025large}. The core challenge of LongDocVQA lies in \emph{multi-hop reasoning}, where intermediate conclusions derived from a single page must be combined with complementary evidence scattered across multiple pages to arrive at the final answer. In high-stakes applications such as financial auditing and medical analysis, only an accurate standalone prediction is insufficient. Users must also be able to inspect how each conclusion is derived from grounded evidence rather than treating model predictions as opaque outputs. This requirement makes explicit evidence composition as important as answer accuracy.

Existing LongDocVQA methods can be broadly categorized into end-to-end MLLMs~\cite{kim2022ocr,liu2026textmonkey,lee2023pix2struct}, retrieval-augmented generation (RAG) pipelines, and agentic frameworks. End-to-end MLLMs implicitly aggregate relevant evidence within hidden representations, making it difficult to inspect how individual evidence contributes to the final prediction. Retrieval-based approaches improve efficiency by selecting relevant pages or regions, yet they still treat retrieved evidence as an unordered context rather than explicitly modeling how evidence is composed during reasoning. Recent document agents expose intermediate actions through iterative tool use, but their reasoning remains trajectory-oriented rather than explicitly representing dependencies among grounded evidence. Although these paradigms differ substantially in how they retrieve and process documents, none explicitly models how grounded evidence is progressively composed into the final answer. Consequently, evidence composition remains implicit, making the reasoning process difficult to verify, supervise, and improve.

Human experts solve long document reasoning by progressively narrowing the search space and organizing grounded evidence into explicit reasoning structures. This naturally follows a hierarchical coarse-to-fine workflow: evidence localization, structured evidence parsing, and evidence composition. More importantly, human reasoning does not merely accumulate evidence; it explicitly models how individual evidence supports intermediate conclusions and how these conclusions jointly lead to the final answer. This observation suggests that evidence composition should be treated as an explicit reasoning object rather than remaining hidden within latent model representations. Motivated by this insight, we propose \textbf{DocTrace}, a hierarchical framework that progressively performs evidence localization, structured parsing, and evidence graph reasoning to explicitly model evidence composition. Specifically, \textbf{DocTrace} progressively performs three stages. It first localizes question-relevant evidence pages from low-resolution document images, thereby reducing the reasoning space. Then it parses the selected pages into structured layout elements with semantic and spatial information. Finally, it constructs an explicit evidence graph over the parsed elements and their visual context, models intermediate reasoning dependencies, and derives the final answer with complete node-level evidence provenance. The overall pipeline is illustrated in Figure~\ref{fig1}.

Learning evidence graph reasoning is challenging because existing LongDocVQA benchmarks provide supervision only for final answers, without annotations for intermediate evidence localization or reasoning graphs. To bridge this gap, we automatically construct supervision consisting of grounded evidence pages and evidence graphs, and use it to jointly initialize evidence localization and graph reasoning via multi-task supervised fine-tuning (SFT). We then refine the model using task-specific Group Relative Policy Optimization (GRPO)~\cite{guo2025deepseek} with dedicated rewards for evidence localization, graph faithfulness and answer correctness. This training paradigm enables \textbf{DocTrace} to construct faithful evidence graphs that inherently provide explicit node-level answer traceability.

Extensive experiments on three long document benchmarks demonstrate that \textbf{DocTrace} consistently outperforms open-source baselines and rivals proprietary models such as GPT-4.1 and Claude-3.7-Sonnet. On the ultra-long benchmark MMLongBench-Doc~\cite{ma2024mmlongbench}, it achieves \textbf{52.9} accuracy, improving the backbone by 14.4 points. It also demonstrates robust performance across varying document lengths. Beyond accuracy, \textbf{DocTrace} provides explicit node-level evidence provenance, making every prediction transparent and verifiable.

Our core contributions are summarized as follows:

\begin{itemize}

\item We propose \textbf{DocTrace}, a \textbf{hierarchical coarse-to-fine framework} for LongDocVQA that progressively performs evidence localization, structured document parsing, and evidence graph reasoning. By explicitly organizing grounded evidence into traceable reasoning graphs, DocTrace enables scalable multi-hop reasoning together with node-level evidence provenance.

\item We develop a dedicated training paradigm for evidence graph reasoning. Specifically, we first jointly initialize evidence localization and graph reasoning through joint SFT on automatically generated training data, and then further optimize both capabilities via task-specific GRPO with dedicated rewards.

\item Extensive experiments on MMLongBench-Doc, LongDocURL, and SlideVQA demonstrate that DocTrace consistently outperforms existing open-source methods, achieving absolute gains of 14.4, 11.3, and 11.7 points over the Qwen3-VL-8B-Instruct backbone on the three benchmarks, respectively, while providing explicit node-level evidence provenance for transparent and traceable reasoning.

\end{itemize}

\begin{figure*}[t]\centering\includegraphics[width=0.9\textwidth]{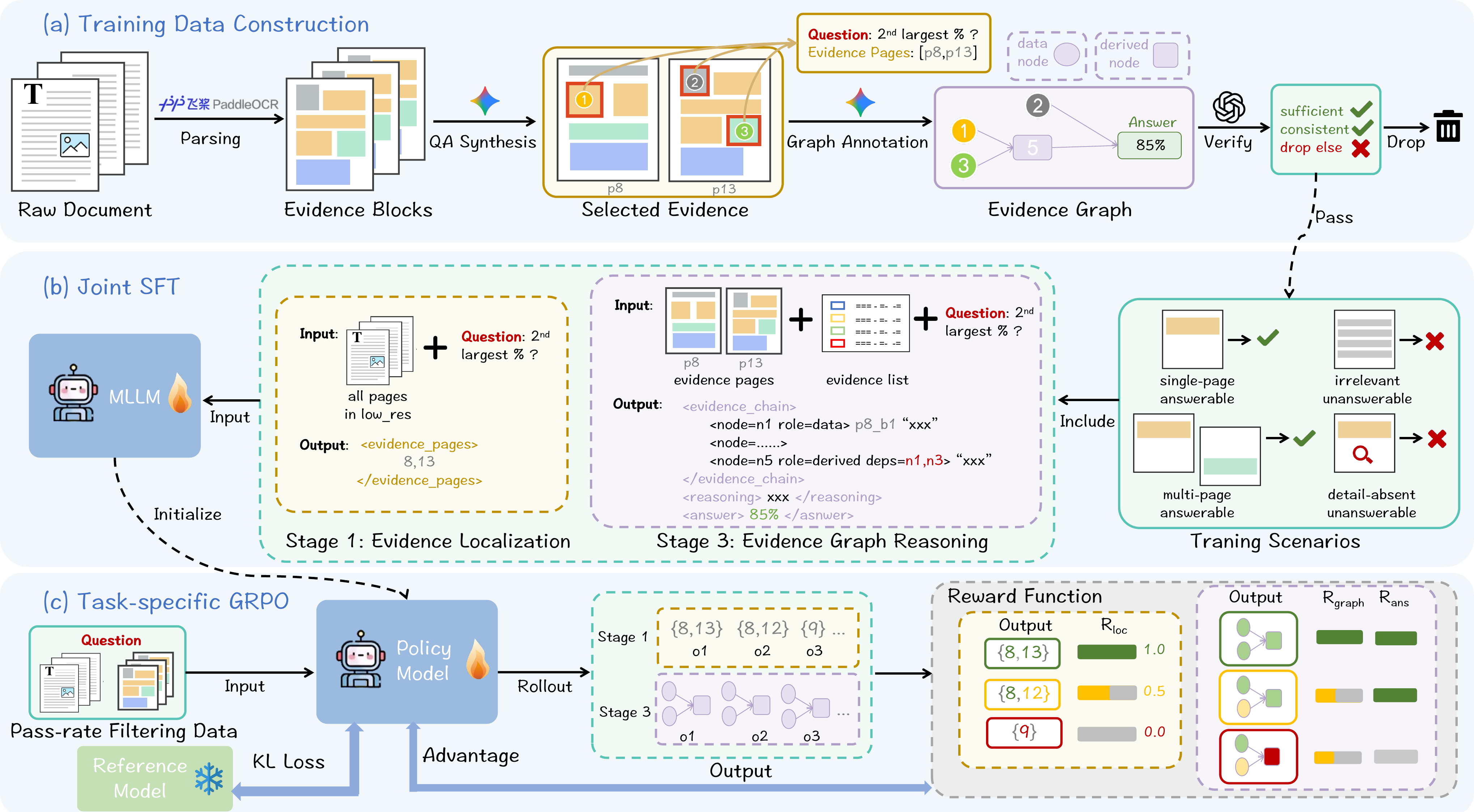}\caption{Overview of our training framework. (a) Training data consturcution using self-collected long docuemnt corpus. (b) Joint SFT, where the model is supervised fine-tuned on high-quality evidence localization and evidence graph reasoning data. (c) Task-specific GRPO, where the policy model is optimized by task-specific reward, including localization reward, graph faithfulness reward and answer correctness reward.}\label{fig2}\end{figure*}

\section{Related Work}

\subsection{Long Document Understanding Dataset}
Long document understanding challenges models to comprehend and reason over lengthy documents, where crucial evidence is often distributed across various pages and modalities. Early datasets, such as {\bf MP-DocVQA}, primarily focus on single-page question answering where the evidence is confined to a specific page. Subsequent works, such as {\bf DUDE}~\cite{van2023document}, {\bf SlideVQA}~\cite{tanaka2023slidevqa}, extend this task to multi-page reasoning and unanswerable questions, emphasizing cross-page localization, multimodal evidence aggregation, and long-horizon reasoning. However, these datasets do not assess documents exceeding 20 pages. More recently, {\bf MMLongBench-Doc} and {\bf LongDocURL}~\cite{deng2025longdocurl}  have scaled the challenge to hundred-page scenarios. Despite this progress, existing training datasets typically provide only final answers together with coarse page-level evidence annotations, offering little supervision for how evidence should be composed during reasoning. Consequently, models must implicitly infer reasoning trajectories from answer alone. To address this limitation, we automatically construct structured evidence supervision with explicit evidence organization and reasoning trajectories, enabling traceable long document reasoning.

\subsection{Methods for Long Document Understanding}

Traditional  {\bf End-to-end } methods (mPLUGDocOwl2~\cite{hu2025mplug}, InternVL3~\cite{zhu2025internvl3}, Qwen3VL~\cite{bai2025qwen3}, Gemini 3) directly process entire documents by leveraging extended context windows or efficient visual token compression. Despite their impressive capability, evidence selection and aggregation remain entirely implicit. {\bf Visual RAG} methods (VisRAG~\cite{yu2025visrag}, SV-RAG~\cite{chen2024sv}, VDocRAG~\cite{tanaka2025vdocrag}) first retrieve the Top-K relevant pages before performing fine-grained reasoning. Recent studies further incorporate layout-aware retrieval, OCR information, and visual representations to improve retrieval quality. However, retrieved evidence is still treated as an unordered collection of pages or chunks, leaving evidence composition to be implicitly inferred by MLLMs and making cross-page reasoning particularly challenging. More recently, {\bf Agent-based } methods (VRAG-RL~\cite{wang2026vrag},  Doc-{\itshape V}$^{*}$~\cite{zheng2026doc}, MM-Doc-R1~\cite{lin2026mm}) have emerged as a promising paradigm, where models actively search, navigate, inspect, and iteratively collect information from long documents through multi-step interactions. Nevertheless, these interaction trajectories do not explicitly model how grounded evidence is composed into the final answer. Overall, existing methods generally represent evidence as implicit context, retrieved pages, or interaction trajectories. Explicit modeling of evidence dependencies remains largely unexplored, limiting both answer traceability and effective supervision for complex cross-page reasoning.

\section{Methodology}

\subsection{Overview}

DocTrace casts LongDocVQA as an evidence graph reasoning problem. Instead of directly predicting answers from long documents, it progressively localizes question-relevant pages, converts them into structured evidence units, and constructs an evidence graph that explicitly models how grounded evidence is composed to derive the final answer. This hierarchical coarse-to-fine design simultaneously reduces the reasoning space while providing node-level evidence provenance. An overview of DocTrace training framework is illustrated in Figure~\ref{fig2}.

\subsection{DocTrace Framework}

\paragraph{Stage~1: Evidence Localization.}

Processing all document pages at native resolution is computationally prohibitive and often exceeds the context capacity of current MLLMs. Therefore, DocTrace first performs coarse evidence localization over uniformly downsampled document pages.

Given the low-resolution document images and question $Q$, DocTrace predicts a set of evidence page indices

\[
\mathcal{P}_{\mathrm{evid}}
=
f_{\theta_{\mathrm{loc}}}
(D_{\mathrm{low}},Q),
\]

where $\mathcal{P}_{\mathrm{evid}}$ denotes the predicted evidence pages. Only these pages are forwarded to subsequent stages, substantially reducing the search space while preserving question-relevant information.

\paragraph{Stage~2: Structured Document Parsing.}

The localized evidence pages are subsequently processed at their native resolution using a document parsing model (PaddleOCR-VL-1.5~\cite{cui2026paddleocr}) to extract fine-grained layout elements. Each element is represented as

\[
b_i=(t_i,\mathbf{x}_i,c_i),
\]

where $t_i$ denotes the semantic type (e.g., text, table, figure, or chart), $\mathbf{x}_i$ is its bounding box, and $c_i$ is the extracted content. Collectively, these parsed elements form an evidence pool

\[
\mathcal{B}=\{b_1,b_2,\ldots,b_M\},
\]

which serves as the atomic evidence units for subsequent graph reasoning.

\paragraph{Stage~3: Evidence Graph Reasoning.}

Given the structured evidence pool, DocTrace explicitly constructs an evidence graph to model how grounded evidence is progressively composed into the final answer. Formally,

\[
G=(V,E),
\]

where each node corresponds to either a grounded evidence block from $\mathcal{B}$ or an intermediate reasoning result, while each directed edge represents a reasoning dependency.

The reasoning process is formulated as

\[
P(A,G\mid\mathcal{B},Q)
=
P(G\mid\mathcal{B},Q)
P(A\mid G,Q),
\]

where the model first constructs the evidence graph and then generates the final answer conditioned on it. Since every reasoning step is explicitly grounded in document evidence, the resulting graph naturally provides node-level evidence provenance for answer verification.

\subsection{Learning DocTrace}
\paragraph{Training Data Construction.}

Existing LongDocVQA benchmarks provide only question-answer pairs, making it impossible to directly supervise evidence localization or evidence graph reasoning. To bridge this supervision gap, we automatically construct structured supervision consisting of evidence pages and evidence graphs.

Specifically, we categorize training samples into four representative scenarios according to answerability and reasoning complexity: (1) single-page answerable, (2) multi-page answerable, (3) irrelevant unanswerable, and (4) detail-absent unanswerable. This taxonomy covers both evidence composition and calibrated refusal behaviors.

For each sample, Gemini 3.1 Pro~\cite{google2026gemini31pro} first generates grounded evidence pages together with the corresponding evidence graph. GPT-5.5~\cite{openai2026gpt55} then independently validates evidence sufficiency and logical consistency. Only verified samples are retained to construct the final supervision corpus for subsequent training.

\paragraph{Joint SFT.}

Using the generated corpus, we jointly optimize evidence localization and evidence graph reasoning through multi-task SFT. Given low-resolution document pages, the model predicts evidence page indices; given localized high-resolution pages and their parsed layout elements, it learns to generate evidence graphs together with the final answers. This stage provides a strong initialization for subsequent GRPO alignment.

\subsubsection{GRPO Alignment}

Although joint SFT provides a strong initialization, it cannot fully optimize evidence localization and graph reasoning. We therefore further align DocTrace using task-specific GRPO objectives: Stage1 optimizes evidence page localization, while Stage3 jointly optimizes evidence graph generation and final answer prediction.

\paragraph{Localization Reward.}

Since page localization is ordinal rather than binary, predictions closer to the ground-truth pages should receive higher rewards than distant ones. We therefore adopt a distance-aware soft $F_{\beta}$ reward that assigns partial credit to nearby predictions while favoring high recall.

Let $\hat{P}$ be the predicted evidence pages parsed from the \texttt{<evidence\_pages>} tag and $G$ the ground-truth evidence pages. Exact matches contribute $h=|\hat{P}\cap G|$, while each unmatched gold page is greedily assigned to its nearest unmatched prediction and receives a distance-dependent soft credit $d(k)$, where $k=|\hat{p}-g|$ is the page-index distance. The soft precision and recall are defined as
\begin{equation}
\mathrm{Prec}=\frac{h+s}{|\hat{P}|},
\qquad
\mathrm{Rec}=\frac{h+s}{|G|},
\end{equation}
where $s$ is the accumulated soft credit. The Stage1 reward is
\begin{equation}
R_{\text{stage1}}
=
\frac{(1+\beta^2)\mathrm{Prec}\cdot\mathrm{Rec}}
{\beta^2\mathrm{Prec}+\mathrm{Rec}},
\end{equation}
where $\beta$ controls the precision--recall trade-off.

To jointly optimize evidence graph generation and answer prediction, the Stage3 reward is defined as

\begin{equation}
R_{\mathrm{stage3}}
=
\lambda
R_{\mathrm{graph}}
+
(1-\lambda)
R_{\mathrm{answer}},
\end{equation}

where $\lambda$ balances evidence graph faithfulness and answer correctness.

\begin{table*}[t]
\centering
\small
\colorlet{primarycol}{gray!6}

\begin{tabular}{lccc>{\columncolor{primarycol}}ccc}
\toprule

\multirow{2}{*}{\textbf{Method}}
&
\multirow{2}{*}{\textbf{Backbone}}
&
\multirow{2}{*}{\textbf{Param.}}
&
\multirow{2}{*}{\textbf{Paradigm}}
&
\cellcolor{primarycol}\textbf{MMLong.}
&
\textbf{LongDoc.}
&
\textbf{SlideVQA}
\\

&
&
&
&
\cellcolor{primarycol}(Acc)
&
(Acc)
&
(F1)
\\

\midrule

\multicolumn{7}{l}{\it\color{gray} Closed Source}\\

Gemini-1.5-Pro
& --
& --
& E2E
& \cellcolor{primarycol}28.2
& 50.9
& --
\\

GPT-4o
& --
& --
& E2E
& \cellcolor{primarycol}42.8
& 64.5
& 65.8
\\

GPT-4.1
& --
& --
& E2E
& \cellcolor{primarycol}45.6
& --
& 74.7
\\

Claude-3.7-Sonnet
& --
& --
& E2E
& \cellcolor{primarycol}33.9
& --
& 76.3
\\

\midrule

\multicolumn{7}{l}{\it\color{gray} Open Source}\\

mPLUG-DocOwl2 \scriptsize\itshape\color{gray}(ACL'25)
& ViT/LLaMA
& 8B
& E2E
& \cellcolor{primarycol}13.4
& 5.3
& 27.8
\\

M3DocRAG \scriptsize\itshape\color{gray}(arXiv'24)
& Qwen2-VL
& 7B
& RAG
& \cellcolor{primarycol}21.0
& 35.1
& 55.7
\\

VisRAG \scriptsize\itshape\color{gray}(ICLR'25)
& MiniCPM-V-2.6
& 8B
& RAG
& \cellcolor{primarycol}18.8
& 41.9
& 52.4
\\

SV-RAG \scriptsize\itshape\color{gray}(ICLR'25)
& InternVL2
& 4B
& RAG
& \cellcolor{primarycol}23.0
& --
& 34.3
\\

VDocRAG \scriptsize\itshape\color{gray}(CVPR'25)
& Phi3-Vision
& 4B
& RAG
& \cellcolor{primarycol}18.4
& 39.8
& 42.0
\\

Docopilot \scriptsize\itshape\color{gray}(CVPR'25)
& InternVL2
& 8B
& E2E
& \cellcolor{primarycol}28.8
& --
& 43.1
\\

InternVL3 \scriptsize\itshape\color{gray}(arXiv'25)
& InternViT/Qwen2.5
& 8B
& E2E
& \cellcolor{primarycol}24.1
& 38.7
& 64.4
\\

VRAG-RL \scriptsize\itshape\color{gray}(NeurIPS'25)
& Qwen2.5-VL
& 7B
& Agent
& \cellcolor{primarycol}26.6
& 44.9
& --
\\

MoLoRAG \scriptsize\itshape\color{gray}(EMNLP'25)
& Qwen2.5-VL
& 7B
& RAG
& \cellcolor{primarycol}41.0
& 51.9
& --
\\

URaG \scriptsize\itshape\color{gray}(AAAI'26)
& Qwen2.5-VL
& 7B
& RAG
& \cellcolor{primarycol}33.8
& 52.2
& --
\\

DocSeeker \scriptsize\itshape\color{gray}(CVPR'26)
& Qwen2.5-VL
& 7B
& E2E
& \cellcolor{primarycol}40.1
& 51.7
& 77.1
\\

Doc-\itshape{V}$^{*}$ \scriptsize\itshape\color{gray}(ACL'26)
& Qwen2.5-VL
& 7B
& Agent
& \cellcolor{primarycol}42.1
& \underline{56.3}
& 77.2
\\

MM-Doc-R1 \scriptsize\itshape\color{gray}(ACL'26)
& Qwen3
& 8B
& Agent
& \cellcolor{primarycol}49.7
& --
& --
\\

\midrule

\rowcolor{gray!8}
\multicolumn{7}{c}{\textit{\textbf{Ours}}}
\\
\cmidrule{1-7}

Qwen3-VL (Baseline)
& Qwen3-VL
& 8B
& E2E
& \cellcolor{primarycol}38.5
& 45.1
& 73.4
\\

\textbf{DocTrace} (SFT)
& Qwen3-VL
& 8B
& Agent
& \cellcolor{primarycol}\underline{50.3}\rlap{\textcolor{red}{$^{\scriptscriptstyle +11.8}$}}
& 53.2\rlap{\textcolor{red}{$^{\scriptscriptstyle +8.1}$}}
& \underline{83.8}\rlap{\textcolor{red}{$^{\scriptscriptstyle +10.4}$}}
\\

\textbf{DocTrace} (GRPO)
& Qwen3-VL
& 8B
& Agent
& \cellcolor{primarycol}\textbf{52.9}\rlap{\textcolor{red}{$^{\scriptscriptstyle +14.4}$}}
& \textbf{56.4}\rlap{\textcolor{red}{$^{\scriptscriptstyle +11.3}$}}
& \textbf{85.1}\rlap{\textcolor{red}{$^{\scriptscriptstyle +11.7}$}}
\\

\bottomrule
\end{tabular}

\caption{Performance comparison on three long document understanding benchmarks: MMLongBench-Doc (Acc), LongDocURL (Acc), and SlideVQA (F1). The \emph{best} and \emph{second-best} results among open-source methods are highlighted in bold and underlined, respectively. Red superscripts denote the \emph{absolute improvement} over Qwen3-VL-8B-Instruct.
}

\label{tab:main}

\end{table*}

\paragraph{Graph Faithfulness Reward.} 
Optimizing only the final answer does not guarantee reasoning faithful to the supporting document evidence. We therefore optimize the generated evidence graph by jointly evaluating evidence grounding, graph completeness, structural validity, and dependency topology: \[ R_{\mathrm{graph}} = w_hR_{\mathrm{hit}} + w_cR_{\mathrm{comp}} + w_sR_{\mathrm{struct}} + w_tR_{\mathrm{topo}}. \] Here, $R_{\mathrm{hit}}$ rewards accurate evidence grounding by maximizing the overlap between predicted and reference evidence nodes. $R_{\mathrm{comp}}$ encourages complete yet compact evidence graphs by discouraging both missing reasoning steps and redundant nodes. $R_{\mathrm{struct}}$ enforces structural validity, including executable DAG constraints and valid evidence references. Finally, $R_{\mathrm{topo}}$ encourages reasoning dependencies that are consistent with the reference graph topology. Together, these rewards promote evidence graphs that are both faithful to the supporting evidence and structurally consistent.

\paragraph{Answer Correctness Reward.}

The answer reward optimizes the correctness of the final prediction:

\[
R_{\mathrm{answer}}
=
\alpha
R_{\mathrm{exact}}
+
(1-\alpha)
R_{\mathrm{llm}},
\]

where $R_{\mathrm{exact}}$ performs exact matching for deterministic answers, while $R_{\mathrm{llm}}$ employs an LLM-based verifier to assess semantic equivalence for answers with flexible surface forms.

For unanswerable questions, the reward encourages abstention when sufficient supporting evidence is unavailable and penalizes hallucinated answers, improving prediction calibration.

The proposed rewards are applied during their corresponding GRPO optimization stages, enabling DocTrace to learn accurate evidence localization, faithful evidence graph reasoning, and reliable answer prediction in a unified reinforcement learning framework.

\section{Experiments}
\subsection{Experimental Setup}

\subsubsection{Datasets.}

For training, we construct a multi-stage training corpus using self-collected long documents (up to 120 pages). It contains synthesized supervision for Stage~1 evidence localization and Stage~3 evidence graph reasoning. See \emph{Appendix} for more details.

For evaluation, we evaluate DocTrace on three long document benchmarks: \textbf{MMLongBench-Doc} (135 documents, avg. 47.5 pages, 1,082 questions; 33.7\% cross-page, 20.9\% unanswerable), \textbf{LongDocURL} (396 documents up to 150 pages, 2,325 QA pairs; 52.9\% multi-page), and \textbf{SlideVQA} (2,215 questions across 20-slide documents; 49.3\% multi-hop/numerical). Notably, MMLongBench-Doc requires whole-document reasoning, whereas LongDocURL uses fixed 30-page windows.

\subsubsection{Evaluation Metrics.}
All methods are evaluated following the \emph{official evaluation protocols} of each benchmark. We report Accuracy on MMLongBench-Doc and LongDocURL, and F1 score on SlideVQA. For fine-grained analysis on MMLongBench-Doc, we additionally report Page F1 and GT page Coverage (Cov.) for evidence localization, alongside Accuracy across Single-page, Multi-page, and Unanswerable question categories for answer generation.

\subsubsection{Implementation Details.}

We adopt \textbf{Qwen3-VL-8B-Instruct} as the backbone and perform full-parameter fine-tuning on 16 NVIDIA A800 GPUs. During SFT, Stage~1 and Stage~3 data are jointly optimized for 3 epochs with a learning rate of $5\times10^{-6}$. During GRPO, Stage~1 and Stage~3 are optimized sequentially for 2 epochs each using a learning rate of $5\times10^{-7}$ and a group size of 8. For Stage~1, the localization reward uses $\beta=2.0$. For Stage~3, the graph faithfulness and answer correctness rewards are weighted by $0.3$ and $0.7$, respectively. The maximum sequence length is 32K during training and 128K during inference. Additional implementation details and hyperparameter settings are provided in \emph{Appendix}.

\subsubsection{Baselines.}

We compare DocTrace against representative baselines spanning three paradigms: (i) {\bf End-to-end} models, including mPLUG-DocOwl2, Docopilot~\cite{duan2025docopilot}, InternVL3,and DocSeeker~\cite{yan2026docseeker}; (ii) {\bf RAG} methods, including M3DocRAG~\cite{cho2024m3docrag}, VisRAG, SV-RAG, VDocRAG, MoLoRAG~\cite{wu2025molorag}, and URaG~\cite{shi2026urag}; and (iii) {\bf Agent-based} approaches, including VRAGRL, Doc-{\itshape V}$^{*}$, and MM-Doc-R1. We additionally include proprietary MLLMs (Gemini-1.5-Pro~\cite{team2024gemini}, GPT-4o~\cite{hurst2024gpt}, GPT-4.1, and Claude-3.7-Sonnet) as reference points, and report the backbone model Qwen3-VL-8B-Instruct for direct comparison. Detailed descriptions of these baseline methods are provided in \emph{Appendix}.

\begin{table}[t]
\centering
\small
\setlength{\tabcolsep}{1mm}
\begin{tabular}{lcccccc}
\toprule
\multirow{2}{*}{\textbf{Train.}}
& \multicolumn{2}{c}{\textbf{Page Loc.}}
& \multicolumn{3}{c}{\textbf{QA Type}}
& \multirow{2}{*}{\textbf{Acc}} \\
\cmidrule(lr){2-3}
\cmidrule(lr){4-6}
& F1 & Cov. & Single & Multi & Unans. & \\
\midrule

Baseline
& -- & --
& 46.9 & 35.3 & 25.8 & 38.5 \\

\midrule

SFT
& 70.8 & 62.4
& 51.7 & 37.7 & \underline{68.0} & 50.3 \\

+ RL (S1)
& \textbf{71.3} & \textbf{65.5}
& \underline{52.6} & \underline{40.8} & 66.0 & \underline{51.5} \\

+ RL (S1+S3)
& \underline{70.8} & \underline{64.6}
& \textbf{53.2} & \textbf{41.1} & \textbf{70.5} & \textbf{52.9} \\

\bottomrule
\end{tabular}
\caption{
Fine-grained performance comparison of the Qwen3-VL-8B-Instruct baseline and DocTrace at different training stages on MMLongBench-Doc. ``F1'' denotes Page F1, and ``Cov.'' denotes ground-truth page coverage.
}
\label{tab:rl_results}
\end{table}

\subsection{Main Results}

\paragraph{Comparison with State-of-the-Art Methods.}

As shown in Table~\ref{tab:main}, DocTrace (GRPO) consistently outperforms both open-source and proprietary models across all three benchmarks. On the challenging MMLongBench-Doc, DocTrace achieves the best accuracy of \textbf{52.9}, outperforming the previous strongest open-source method MM-Doc-R1 (49.7) by 3.2 points and the proprietary model GPT-4.1 (45.6) by 7.3 points. On LongDocURL, DocTrace achieves the highest accuracy of \textbf{56.4}, slightly surpassing Doc-{\itshape V}$^{*}$ (56.3). On SlideVQA, DocTrace establishes a new state of the art with \textbf{85.1} F1, exceeding Doc-{\itshape V}$^{*}$ by 7.9 points. These results show that explicit evidence graph reasoning consistently improves long-document understanding across diverse benchmarks.

\paragraph{Effectiveness of the Training Paradigm.}

We further evaluate the effectiveness of our two-stage training paradigm, which combines joint SFT with task-specific GRPO. As shown in Table~\ref{tab:main}, joint SFT substantially improves all three benchmarks over the Qwen3-VL-8B backbone, yielding gains of 11.8, 8.1, and 10.4 points on MMLongBench-Doc, LongDocURL, and SlideVQA, respectively. Building on this strong initialization, task-specific GRPO provides consistent additional improvements, achieving the best overall performance of 52.9, 56.4, and 85.1 on the three benchmarks.

\paragraph{Analysis of Task-Specific GRPO.}

To better understand the complementary effects of task-specific GRPO, Table~\ref{tab:rl_results} presents a fine-grained analysis on MMLongBench-Doc.

Applying GRPO to Stage~1 primarily improves evidence localization, increasing Page F1 from 70.8 to 71.3 and page coverage from 62.4 to 65.5. Better evidence retrieval consistently benefits both single-page (51.7$\rightarrow$52.6) and multi-page (37.7$\rightarrow$40.8) questions. Meanwhile, the accuracy on unanswerable questions drops slightly (68.0$\rightarrow$66.0), suggesting that improved evidence recall also encourages more aggressive answering when supporting evidence is insufficient.

Applying GRPO to Stage~3 further improves answer quality through more effective evidence aggregation and reasoning. Although retrieval metrics decrease slightly due to optimization trade-offs, answer accuracy continues to improve on single-page (52.6$\rightarrow$53.2), multi-page (40.8$\rightarrow$41.1), and especially unanswerable questions (66.0$\rightarrow$70.5), resulting in the best overall accuracy of 52.9.

Together, these observations highlight the complementary roles of the two optimization stages: Stage~1 strengthens evidence acquisition, whereas Stage~3 improves evidence utilization and answer reliability.

\subsection{Analysis of Long-Document Scalability}

To evaluate the scalability of DocTrace on increasingly long documents, we conduct a controlled experiment on a subset of MMLongBench-Doc containing documents longer than 60 pages. For each target length $W$, documents are truncated or padded with answer-free pages while keeping the questions unchanged, isolating the effect of document length.

As shown in Figure~3(a), the performance of the Qwen3-VL-8B baseline deteriorates rapidly as document length increases, with accuracy dropping from 49.5 to 34.0. In contrast, DocTrace consistently outperforms the baseline and maintains much more stable performance across different document lengths.

Figure~3(b) provides further insight into this degradation. Although evidence retrieval coverage gradually decreases from 70.6 to 50.2 as documents become longer, the answer accuracy conditioned on successful evidence localization remains largely unchanged. This suggests that DocTrace's evidence graph reasoning remains robust to increasing document length, and that further scalability improvements mainly depend on stronger evidence localization.

\begin{figure}[h]
      \centering
      \includegraphics[width=0.9\columnwidth]{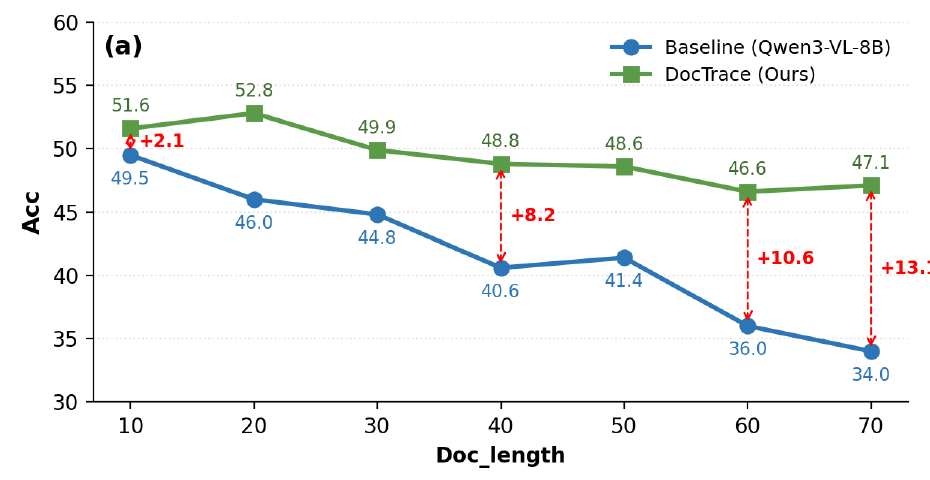}
      \includegraphics[width=0.9\columnwidth]{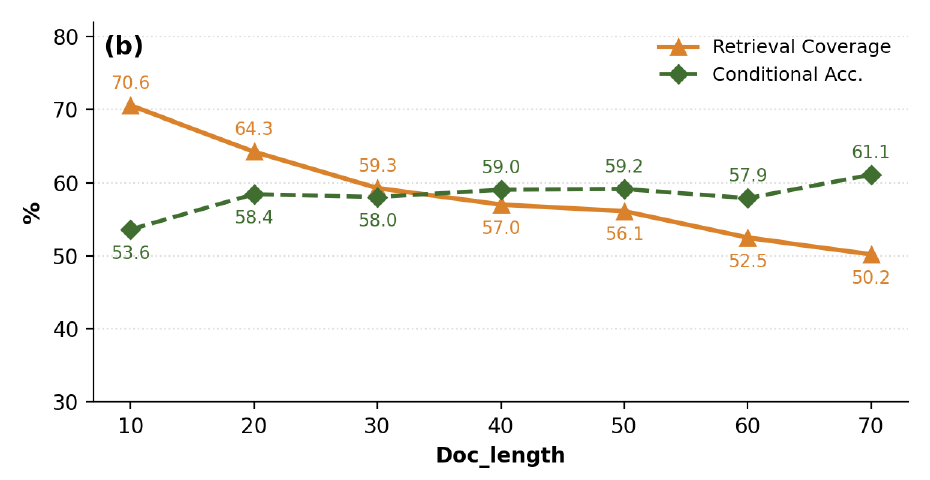}
      \caption{Performance comparison on different document length. (a) Acc; (b) Retrieval Cov. vs.\ cond Acc.}
      \label{fig3}
\end{figure}

\subsection{Traceability Analysis of Evidence Graph Reasoning}

To assess the reliability of DocTrace evidence graph reasoning, we perform both qualitative visualizations and quantitative evaluations on MMLongBench-Doc. More details are provided in \emph{Appendix}.

As illustrated in Figure~\ref{fig:showcase}, DocTrace constructs an explicit evidence graph that links supporting evidence across multiple document pages through intermediate reasoning steps. In this example, the model associates demographic information on Page~8 with the corresponding Wi-Fi promotion on Page~13, producing a transparent reasoning path from grounded evidence to the final answer.

We further evaluate the traceability of the generated evidence graphs from three perspectives: provenance integrity, evidence localization, and causal faithfulness (Table~\ref{tab:traceability}). DocTrace achieves 99.5\% evidence grounding accuracy and 99.6\% graph integrity, indicating that the generated reasoning traces are consistently anchored to valid document evidence. The evidence localization F1 reaches 72.5\%, demonstrating effective retrieval of supporting evidence from long documents. Furthermore, counterfactual evaluation shows that masking cited evidence flips 82.8\% of originally correct predictions, whereas masking uncited evidence changes only 9.6\%. This large gap indicates that the generated evidence graphs capture the evidence that genuinely supports the model's predictions.

\begin{figure}[h]
\centering
\includegraphics[width=0.9\linewidth]{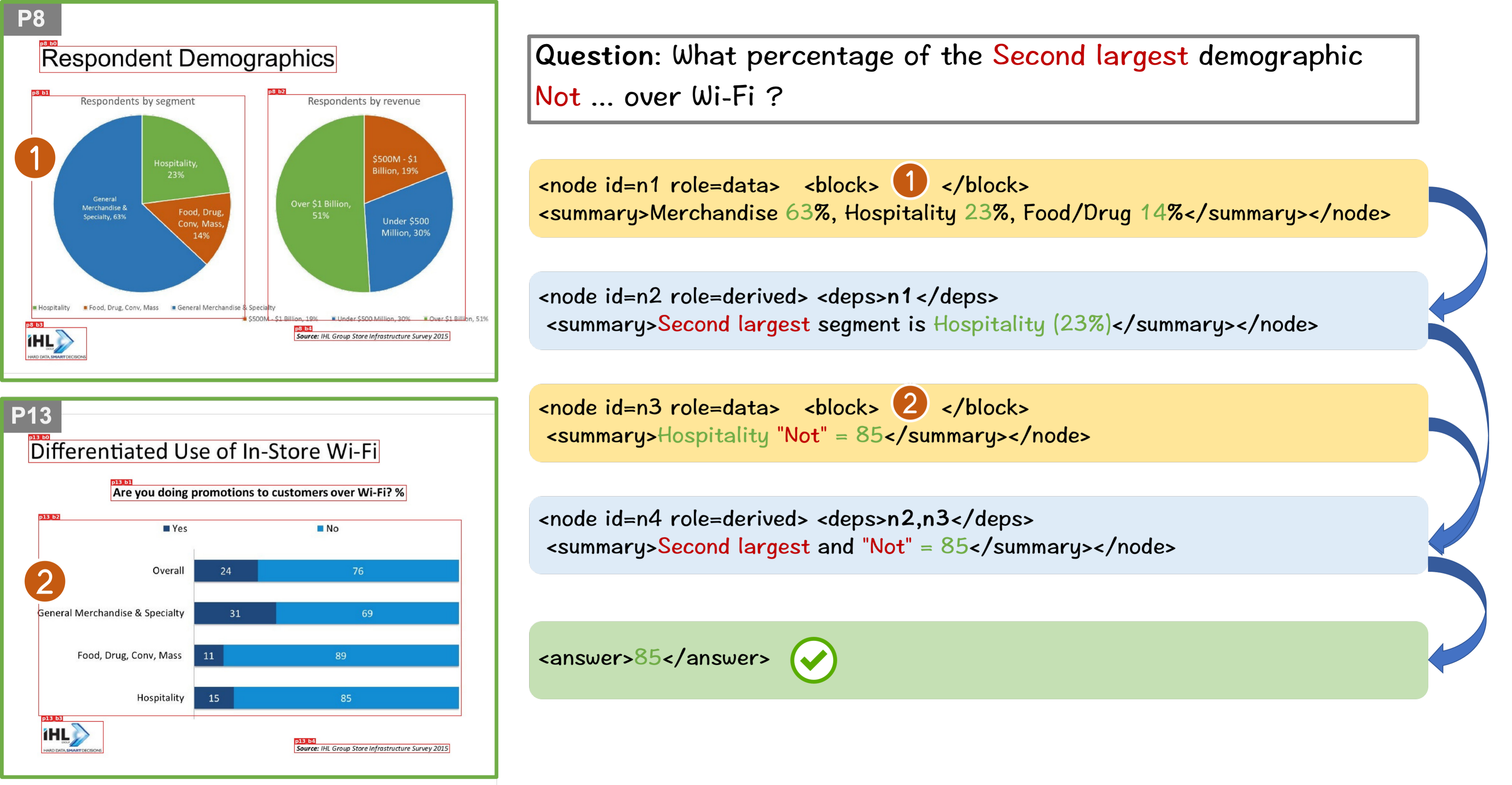}
\caption{
Qualitative visualization of DocTrace evidence graph reasoning for a multi-page reasoning task.
}
\label{fig:showcase}
\end{figure}

\begin{table}[h]
\centering
\small
\begin{tabular}{lcc}
\toprule
\textbf{Traceability Metric} & \textbf{Score} & \textbf{Verifier} \\
\midrule

\multicolumn{3}{l}{\textit{Provenance Integrity}} \\
\quad Evidence Grounding & 99.5 & Rule \\
\quad Evidence Graph Integrity & 99.6 & Rule \\

\midrule

\multicolumn{3}{l}{\textit{Evidence Localization}} \\
\quad Evidence Localization F1 & 72.5 & GT \\

\midrule

\multicolumn{3}{l}{\textit{Causal Faithfulness}} \\
\quad Evidence Necessity & 82.8 & C.F. \\
\quad Evidence Specificity & 9.6 & C.F. \\

\bottomrule
\end{tabular}

\caption{
Quantitative evaluation of DocTrace evidence graph traceability on MMLongBench-Doc.
Scores are percentages. C.F. denotes counterfactual evaluation by masking
evidence and re-evaluating the same model.
}
\label{tab:traceability}
\end{table}


\subsection{Ablation Study}

We conduct ablation studies on MMLongBench-Doc based on the SFT model to evaluate the contribution of the structural parsing and evidence graph reasoning components in DocTrace. The results are summarized in Table~\ref{tab:ablation_results}.

\paragraph{Effect of Core Components.}
Removing either structural parsing or evidence graph reasoning consistently degrades the overall performance, reducing the accuracy from 50.3 to 46.6 and 47.2, respectively. Both variants suffer substantial performance drops on multi-page questions, demonstrating that structured representations and explicit evidence graph reasoning are essential for integrating dispersed evidence across long documents. Compared with removing structural parsing, removing evidence graph reasoning preserves stronger single-page performance (47.1 vs. 44.6), suggesting that parsed structures remain effective for local evidence understanding, while graph reasoning primarily contributes to multi-hop reasoning over cross-page evidence. Meanwhile, both variants achieve higher unanswerable accuracy (74.2 and 74.5), indicating a more conservative prediction behavior that improves unanswerable detection but compromises answerable question solving.

\paragraph{Comparison with Vanilla CoT.}
Replacing evidence graph reasoning with vanilla CoT~\cite{wei2022chain} achieves 47.8 accuracy. Although it slightly improves single-page QA (53.2), it performs worse on multi-page questions (34.8) and unanswerable questions (58.2). This suggests that vanilla linear reasoning can handle local evidence aggregation but struggles to preserve structured evidence dependencies and effectively handle complex long document reasoning.

\begin{table}[h]
\centering
\small

\begin{tabular}{lcccc}
\toprule
\multirow{2}{*}{\textbf{Setting}}
& \multicolumn{3}{c}{\textbf{QA Type}}
& \multirow{2}{*}{\textbf{Acc}} \\
\cmidrule(lr){2-4}
& Single & Multi & Unans. & \\
\midrule

DocTrace (SFT)
& 51.7
& \textbf{37.7}
& 68.0
& \textbf{50.3} \\

\midrule

w/o Structural Parsing
& 44.6
& 30.6
& \textbf{74.2}
& 46.6 \\

w/o Graph Reasoning
& 47.1
& 30.2
& \textbf{74.5}
& 47.2 \\

w/ Vanilla CoT
& \textbf{53.2}
& 34.8
& 58.2
& 47.8 \\

\bottomrule
\end{tabular}

\caption{
Ablation study of structural parsing and evidence graph reasoning on MMLongBench-Doc.
}
\label{tab:ablation_results}
\end{table}

\section{Conclusion}

In this paper, we cast LongDocVQA as an explicit evidence graph reasoning problem and propose DocTrace, a hierarchical framework that progressively performs evidence localization, structured document parsing, and evidence graph reasoning. We further develop a two-stage training framework combining SFT and task-specific GRPO to improve evidence acquisition and reasoning capabilities. Experimental results demonstrate that DocTrace achieves strong and robust performance across multiple long-document benchmarks. Beyond competitive performance, DocTrace provides explicit node-level evidence provenance, enabling transparent and verifiable reasoning for long document understanding.

\bibliography{aaai2027}

\clearpage

\appendix

\twocolumn[
\centering
{\LARGE\bfseries Appendix\par}
\vspace{2.0em}
]

\section{Training Data Construction}

\subsection{Overview}

This section describes the construction of the supervision corpus used for both supervised fine-tuning (SFT) and Group Relative Policy Optimization (GRPO). Starting from publicly available LongDocVQA benchmarks and self-collected long documents, we automatically generate evidence page annotations and structured evidence graphs through a teacher--verifier pipeline. The resulting supervision corpus provides explicit annotations for evidence localization, evidence graph reasoning, and answer prediction.

\subsection{Data Source}

The training corpus is automatically constructed from three publicly available LongDocVQA benchmarks, namely MPDocVQA, DUDE, and SlideVQA. We further augment the training corpus with a self-collected set of long documents, primarily consisting of publicly available arXiv papers and technical reports. While the documents in the public LongDocVQA benchmarks are generally shorter than 20 pages, the self-collected corpus includes substantially longer documents, with document lengths of up to 120 pages. We use the original documents together with their question--answer annotations for the public benchmarks as the starting point for automatic evidence graph generation.

\begin{itemize}

\item \textbf{MPDocVQA}
contains scanned multi-page documents paired with visual question answering annotations. Most questions require locating fine-grained textual evidence from one or several pages, making it well suited for supervising evidence localization and document grounding.

\item \textbf{DUDE}
extends document understanding to more challenging multi-page reasoning scenarios. Its documents exhibit diverse layouts, including forms, reports, tables, and visually rich pages, requiring models to retrieve and integrate evidence distributed across multiple pages.

\item \textbf{SlideVQA}
focuses on presentation slides containing figures, charts, diagrams, bullet lists, and sparse textual content. Since slide documents often follow non-linear visual layouts rather than continuous textual flow, they provide complementary supervision for multimodal reasoning over heterogeneous document elements.

\item \textbf{Self-Collected Long Documents}
consist primarily of publicly available arXiv papers and technical reports. Compared with the public LongDocVQA benchmarks, these documents are substantially longer (up to 120 pages) and exhibit rich cross-page dependencies, providing additional supervision for long-range evidence localization and reasoning beyond the length range covered by existing benchmarks.

\end{itemize}

\subsection{Automatic Evidence Graph Generation Pipeline}

Given a document and a question, a teacher MLLM first identifies the evidence pages required to answer the question. It then performs fine-grained reasoning over the selected pages and constructs an explicit evidence graph describing grounded evidence nodes, intermediate reasoning nodes, dependency relations, and the final answer.

Each generated evidence graph consists of four components:

\begin{itemize}

\item \textbf{Evidence Pages.}
The document page indices required for answering the question.

\item \textbf{Evidence Nodes.}
Grounded document elements supporting the reasoning process. Each node corresponds to one parsed document block.

\item \textbf{Derived Nodes.}
Intermediate reasoning results obtained by composing one or multiple evidence nodes.

\item \textbf{Dependency Edges.}
Directed edges that describe how grounded evidence is progressively combined into higher-level reasoning results until reaching the final answer.

\end{itemize}

The resulting graph forms an executable directed acyclic graph (DAG), where every reasoning step remains explicitly grounded in document evidence.

To ensure annotation quality, every generated sample is subsequently validated by an independent verifier MLLM before being included in the final supervision corpus.

\begin{figure*}[t]
\centering
\includegraphics[width=0.9\textwidth]{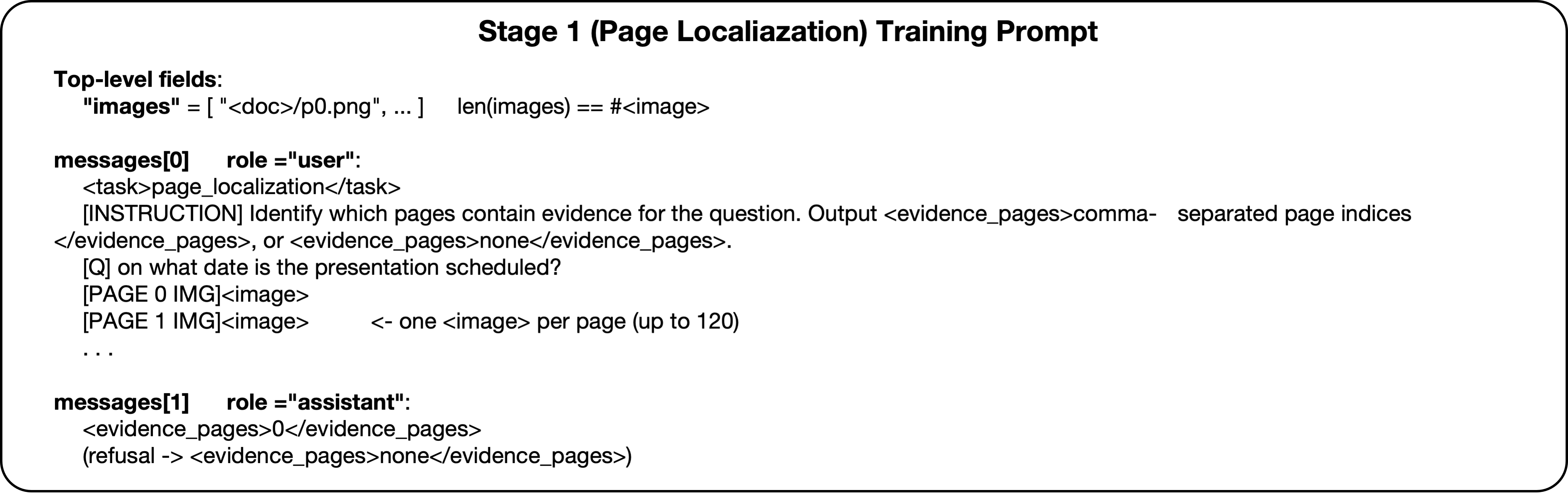}
\\[1em]
\includegraphics[width=0.9\textwidth]{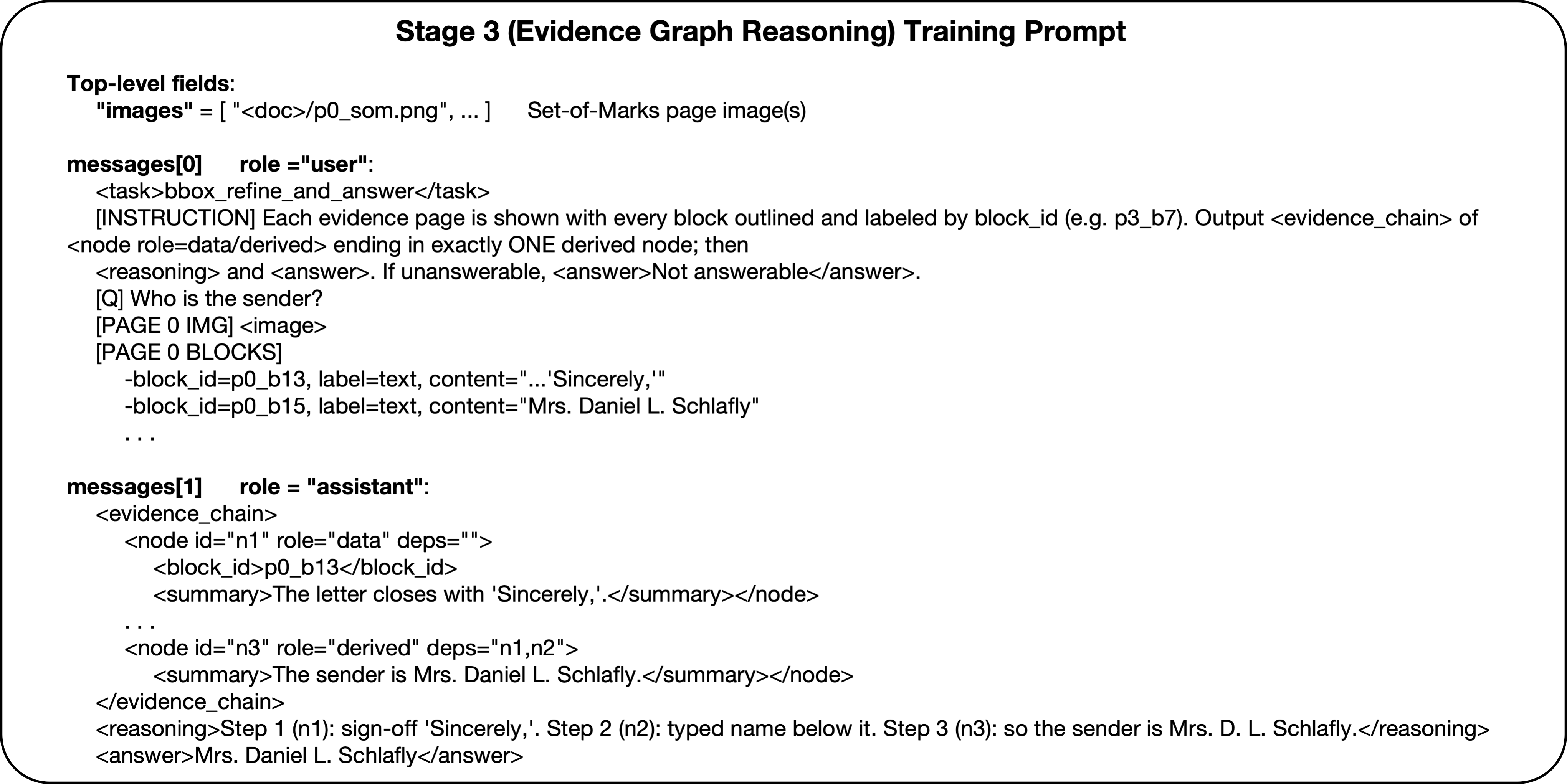}
\caption{Top: Prompts used for Stage 1 Page Localization Training. Bottom: Prompts used for Stage 3 Evidence Graph Reasoning Training.}
\label{fig:Stage1_Stage3_prompt}
\end{figure*}

\subsection{Data Verification}

Since evidence graphs are generated automatically, annotation quality is controlled through an independent verification stage.

For each generated sample, the verifier evaluates three complementary aspects.

\paragraph{Evidence Sufficiency.}
The verifier checks whether the predicted evidence pages contain sufficient information to answer the given question. Samples requiring additional unseen evidence are discarded.

\paragraph{Logical Consistency.}
The verifier examines whether every reasoning step is logically supported by its predecessors and whether the dependency graph forms a coherent reasoning process without missing or contradictory intermediate nodes.

\paragraph{Answer Correctness.}
Finally, the verifier independently derives the answer from the generated evidence graph and compares it with the reference answer. Samples with inconsistent predictions are removed.

Only samples satisfying all verification criteria are retained for subsequent SFT and GRPO training.

\subsection{Data Statistics}

Tables~\ref{tab:data_statistics} and~\ref{tab:doc_statistics} summarize the statistics of the constructed supervision corpus.

We first report the composition of answerable and unanswerable questions, together with the distribution of different reasoning scenarios. We further present statistics of evidence pages, including the average number of document pages and evidence pages per sample.

The resulting supervision corpus covers both answerable and unanswerable settings, with diverse evidence localization complexity ranging from single-page retrieval to cross-page evidence composition.

\begin{table}[h]
\centering
\begin{tabular}{lcccc}
\toprule
\textbf{Task}
& \textbf{Ans.}
& \textbf{Irr.}
& \textbf{D.A.}
& \textbf{Total} \\
\midrule

Stage~1
&
\makecell[c]{17,102\\[-1mm]
{\scriptsize(71.2\% SP)}}
&
1,001
&
--
&
18,103
\\

Stage~3
&
\makecell[c]{15,998\\[-1mm]
{\scriptsize(73.4\% SP)}}
&
--
&
2,718
&
18,716
\\

\midrule

\textbf{Total}
&
\textbf{33,100}
&
\textbf{1,001}
&
\textbf{2,718}
&
\textbf{36,819}
\\

\bottomrule
\end{tabular}

\caption{
Composition of the two-stage training corpus.
Irr. denotes irrelevant-page refusals, D.A. denotes detail-absent refusals and SP indicates single-page questions.
}
\label{tab:data_statistics}

\end{table}

\begin{table}[h]
\centering

\begin{tabular}{lc}
\toprule
\textbf{Metric} & \textbf{Value} \\
\midrule
\multicolumn{2}{c}{\textit{Document Statistics}}\\
\midrule
Avg. document pages & 26.1 \\
Max. document pages & 120 \\
\midrule
\multicolumn{2}{c}{\textit{Evidence Statistics}}\\
\midrule
Avg. evidence pages / QA & 1.4 \\
Max. evidence pages / QA & 29 \\
\bottomrule
\end{tabular}

\caption{
Document and evidence statistics of the automatically constructed supervision corpus.
Evidence pages denote the pages required to answer a question.
}
\label{tab:doc_statistics}

\end{table}

\begin{table}[t]
\centering

\begin{tabular}{lc}
\toprule
  \textbf{Parameter} & \textbf{Value} \\
  \midrule
  \multicolumn{2}{c}{\textit{Shared}} \\
  \midrule
  Backbone              & Qwen3-VL-8B-Instruct \\
  Training Precision    & BF16 \\
  Optimizer             & AdamW \\
  LR Scheduler          & Cosine \\
  Warmup Ratio          & 0.05 \\
  Gradient Clipping     & 1.0 \\
  DeepSpeed             & ZeRO-2 \\
  FlashAttention        & Enabled \\
  Gradient Checkpointing& Enabled \\
  Training GPUs         & 16 $\times$ NVIDIA A800 (80\,GB) \\
  \midrule
  \multicolumn{2}{c}{\textit{SFT}} \\
  \midrule
  Learning Rate         & $5\times10^{-6}$ \\
  Epochs                & 3 \\
  Max Sequence Length   & 32K (with sequence packing) \\
  Effective Batch Size  & 8 \\
  Liger Kernel          & Enabled \\
  \midrule
  \multicolumn{2}{c}{\textit{GRPO}} \\
  \midrule
  Learning Rate         & $5\times10^{-7}$ \\
  Epochs                & 2 \\
  Group Size ($G$)      & 8 \\
  Rollout Temperature   & 0.8 \\
  Rollout Top-$p$       & 0.95 \\
  Max Prompt Length     & 32K \\
  Clip Range ($\epsilon_{\text{low}}/\epsilon_{\text{high}}$) & 0.2 / 0.28 \\
  Effective Batch Size  & 24 \\
\bottomrule
\end{tabular}

\caption{Training hyperparameters of DocTrace.}
\label{tab:hyperparameters}
\end{table}

\section{Additional Experimental Details}

\subsection{Implementation Details}

To facilitate reproducibility, we provide additional implementation details beyond those reported in the main paper.

DocTrace is built upon Qwen3-VL-8B-Instruct and uses full-parameter optimization throughout training. Unless otherwise specified, all experiments are conducted on 16 NVIDIA A800 GPUs with BF16 mixed-precision training. FlashAttention-2 and gradient checkpointing are enabled for memory-efficient long-context training.

Following the hierarchical design of DocTrace, training consists of one SFT stage followed by two task-specific GRPO stages. During SFT, Stage~1 evidence localization and Stage~3 evidence graph reasoning are jointly optimized with a unified multi-task objective. Stage~1 uses document images at a fixed resolution of $512$, while Stage~3 processes the selected evidence pages at their native resolution of $1568$ for fine-grained evidence parsing and graph reasoning. Figure illustrates the prompts used for DocTrace training.

After SFT, reinforcement learning is performed sequentially. Stage~1 is first optimized with the localization reward, and the resulting checkpoint is then used to initialize Stage~3 GRPO, where graph faithfulness and answer correctness are jointly optimized. The same resolution settings are used during GRPO: $512$ for Stage~1 and native $1568$ for Stage~3. This sequential strategy allows graph reasoning to be optimized on top of improved evidence localization.

Unless otherwise specified, all experimental results reported in the main paper are obtained from the final Stage~3 GRPO checkpoint.

\begin{table}[t]
\centering

\begin{tabular}{lc}
\toprule
\textbf{Parameter} & \textbf{Value} \\
\midrule
Localization reward $\beta$ & 2.0 \\
Stage~3 reward $\lambda$ & 0.3 \\
Answer reward $\alpha$ & 0.5 \\
$w_h$ (Hit) & 0.55 \\
$w_c$ (Completeness) & 0.20 \\
$w_s$ (Structure) & 0.15 \\
$w_t$ (Topology) & 0.10 \\
\bottomrule
\end{tabular}

\caption{Reward hyperparameters used during GRPO.}
\label{tab:reward_hyper}

\end{table}

\subsection{Training Hyperparameters}

Table~\ref{tab:hyperparameters} summarizes the optimization hyperparameters used throughout training. The reward hyperparameters for the Stage~1 localization objective and the Stage~3 evidence graph optimization objective are reported separately in Table~\ref{tab:reward_hyper}.

\subsection{GRPO Training Candidate Construction}

Directly applying GRPO to all training samples provides limited learning signals, since samples that are always solved correctly or consistently fail produce nearly zero relative advantage within each rollout group. We therefore construct informative GRPO training candidates through an offline rollout procedure.

\paragraph{Stage~1 Candidate Selection.}

Starting from the joint SFT checkpoint, we perform eight independent rollouts for every Stage~1 training sample. Samples that are solved correctly in all eight rollouts or fail in all rollouts are discarded, since they contribute little useful optimization signal. Only samples with intermediate success rates (i.e., 1/8 to 7/8 correct predictions) are retained as Stage~1 GRPO training candidates.

\paragraph{Stage~3 Candidate Selection.}

After Stage~1 GRPO converges, the resulting checkpoint is used to generate rollouts on the Stage~3 supervision corpus following the same procedure. Again, only samples with intermediate success rates are retained for reinforcement learning.

To further improve training efficiency, we slightly adjust the sampling ratio of different reasoning scenarios according to the performance of the SFT model. Categories that are already well mastered (e.g., single-page answerable questions) are moderately downsampled, while more challenging categories, including multi-page answerable questions and detail-absent unanswerable questions, are upsampled. This curriculum-style sampling strategy allocates more optimization effort to difficult reasoning behaviors that benefit most from reinforcement learning.

\section{Additional Ablation Study}
\subsection{The Impact of Set-of-Marks}

We conduct an additional ablation experiment to study the impact of the Set-of-Marks (SoM) representation in Stage~3. The default SoM setting provides visually grounded layout information by rendering block boundaries and identifiers on evidence pages, together with cropped visual regions for chart/figure blocks.

We compare it with a textual bounding-box variant, where the original clean page image is used without visual annotations. Instead, each block is represented by its block id, semantic label, and normalized bounding-box coordinates in the prompt, while chart/figure crops are removed. All other components, including training data, supervision targets, backbone, and optimization settings, are kept identical to the SoM baseline.

\begin{table}[h]
\centering
\begin{tabular}{lcccc}
\toprule
\multirow{2}{*}{\textbf{Setting}}
& \multicolumn{3}{c}{\textbf{QA Type}}
& \multirow{2}{*}{\textbf{Acc}} \\
\cmidrule(lr){2-4}
& Single & Multi & Unans. & \\
\midrule

DocTrace (SFT)
& \textbf{51.7}
& \textbf{37.7}
& \textbf{68.0}
& \textbf{50.3} \\

\midrule

w/o SoM
& 50.4
& 32.5
& 67.6
& 48.1 \\

\midrule

$\Delta$
& -1.3
& -5.2
& -0.4
& -2.2 \\

\bottomrule
\end{tabular}
\caption{
Ablation study of SoM on MMLongBench-Doc.
}
\label{tab:ablation_results}
\end{table}

The results are summarized in Table~\ref{tab:ablation_results}. The textual bounding-box variant consistently underperforms SoM, with an overall accuracy drop of 2.2 points (48.1 vs.\ 50.3), demonstrating the benefit of explicit visual grounding for fine-grained evidence localization and document reasoning. The performance gap is primarily observed on multi-page questions, where the textual variant drops by 5.2 points (32.5 vs.\ 37.7). In contrast, single-page questions show only a minor degradation (50.4 vs.\ 51.7, $-1.3$), and unanswerable questions remain nearly unchanged (67.6 vs.\ 68.0, $-0.4$), suggesting that SoM mainly benefits cross-page evidence integration.

We attribute the larger gain on multi-page questions to the stronger spatial grounding provided by SoM. By rendering layout blocks and their block ids directly on the page, SoM establishes an explicit visual correspondence between reasoning steps and document regions. In comparison, the textual variant requires the model to recover the same spatial relationships from normalized coordinates, which is more challenging for cross-page reasoning.

\section{Detailed Traceability Analysis}

This section provides the evaluation details for the traceability results reported in Table~3 of the main paper. We evaluate traceability from three complementary perspectives: provenance integrity, evidence localization, and causal faithfulness. Only samples that successfully reach Stage~3 and produce a valid evidence chain are included in graph-based evaluations.

\subsection{Provenance Integrity}

We verify the structural validity and grounding of each generated evidence graph using deterministic rules without model calls. Specifically, we check whether (1) all cited block IDs exist in the Stage~2 layout inventory, (2) all dependency references are valid and every derived node has parents, (3) the graph is acyclic, and (4) the final answer node is connected to grounded evidence.

Among 1,049 valid chains containing 3,122 evidence nodes, 99.5\% of cited block IDs are successfully grounded. The graph well-formedness, acyclicity, and answer connectivity rates are 99.6\%, 100.0\%, and 100.0\%, respectively, indicating that the generated traces are structurally reliable and well grounded.

\subsection{Evidence Localization}

We evaluate whether the evidence used by the reasoning trace corresponds to human-annotated evidence pages. For each answerable sample with non-empty evidence annotations, we collect the pages referenced by evidence nodes and compute macro-averaged precision, recall, and $F_1$ against the ground-truth evidence pages. We additionally compare the result with the Stage~1 retrieved page set.

On 813 answerable samples, the reasoning trace achieves $P/R/F_1=75.9/71.7/72.5$, compared with an $F_1$ score of 72.2 for Stage~1 retrieval. This shows that the generated reasoning traces effectively consume the retrieved evidence.

\subsection{Causal Faithfulness}

We further test whether the cited evidence is functionally necessary for the final prediction through counterfactual masking. For each sample, we compare two conditions:

\begin{itemize}
\item \textbf{Cited masking}: mask the blocks referenced by the evidence graph.
\item \textbf{Uncited masking}: mask the same number of grounded but uncited blocks as a placebo control.
\end{itemize}

All other inputs remain unchanged. Among originally correct answerable samples with grounded evidence ($n=363$), 319 samples satisfy the cited-masking control condition. Masking cited evidence changes the answer in 82.8\% of cases (264/319), whereas masking uncited evidence changes the answer in only 9.6\% of cases (30/313). The substantial gap indicates that the evidence identified by the graph is functionally important for the final prediction.

\subsection{Limitation}

This evaluation measures behavioral dependence through input perturbation. It does not establish that the model internally follows the explicit evidence graph; rather, it shows that the cited evidence is functionally important for the model's prediction.

\section{Efficiency Analysis}

We compare the inference efficiency of DocTrace with the end-to-end Qwen3-VL-8B baseline on the full MMLongBench-Doc benchmark. Both methods are evaluated on identical hardware with batch size 1. The end-to-end baseline processes all document pages at 1024\,px in a single forward pass, whereas DocTrace first localizes evidence using a Stage~1 scan at 512\,px, parses the retrieved pages in Stage~2, and performs fine-grained reasoning over the retrieved evidence pages at their native resolution of 1568\,px in Stage~3.

Table~\ref{tab:efficiency} shows that DocTrace is both more accurate and more efficient than the end-to-end baseline. It improves accuracy from 38.5 to 52.9 (+14.4 points), while reducing end-to-end latency by $2.1\times$ (15.04\,s to 7.33\,s) and prefill tokens by $2.3\times$ (35.5K to 15.5K). The efficiency gain comes from the hierarchical design: Stage~1 performs a lightweight scan over all document pages using compact 512\,px inputs, while Stage~3 applies high-resolution reasoning only to the retrieved evidence pages. As a result, DocTrace concentrates computation on pages where fine-grained visual reasoning is required, rather than processing every page at high resolution.

\begin{table}[h]
\centering
\small
\begin{tabular}{lccc}
\toprule
Method & ACC & E2E latency (s) & Prefill tok. \\
\midrule
Qwen3-VL-8B (E2E) & 38.5 & 15.04 & 35,543 \\
\midrule
\textbf{DocTrace} & \textbf{52.9} & \textbf{7.33} & \textbf{15,481} \\
\quad Stage~1 (512\,px) & -- & 1.67 & 9,269 \\
\quad Stage~2 (OCR) & -- & 0.28 & -- \\
\quad Stage~3 (1568\,px) & -- & 5.38 & 6,212 \\
\bottomrule
\end{tabular}
\caption{
Efficiency comparison on MMLongBench-Doc. Prefill tokens include both visual and text tokens.
}
\label{tab:efficiency}
\end{table}

\section{Motivation for Hierarchical Inference}

DocTrace adopts a coarse-to-fine inference strategy: Stage~1 performs evidence localization over the entire document using low-resolution page images (512\,px), while Stage~3 revisits only the retrieved evidence pages at high resolution (1568\,px) for fine-grained reasoning. This design is motivated by a fundamental limitation of end-to end inference: processing every page at high resolution rapidly exhausts the visual context budget of current MLLMs. To quantify this limitation, we analyze the context-overflow rate under different page resolutions and document lengths.

For a page rendered at long-edge resolution $R$, Qwen3-VL produces $t_R(p)=\lfloor W_R(p)H_R(p)/1024\rfloor+8$ visual tokens, where both image dimensions are rounded to multiples of 32. A document overflows whenever the total visual and textual tokens exceed the available context window:

\begin{equation}
\sum_p t_R(p)+T_{\text{text}}
>
C-M_{\text{gen}}-S,
\label{eq:overflow}
\end{equation}

where $C\in\{256\mathrm{K},128\mathrm{K}\}$ is the context window, $M_{\text{gen}}=2048$ reserves generation tokens, $S=64$ is a safety margin, and $T_{\text{text}}=512$ denotes the textual prompt budget. The token accounting is identical to that used during inference. Table~\ref{tab:overflow} reports the resulting overflow rates on MMLongBench-Doc.

Table~\ref{tab:overflow} reveals that the overflow rate increases sharply with page resolution. Under the 256K setting, 2048\,px already overflows for 79.0\% of 80--120-page documents and for all documents longer than 120 pages, while 1568\,px also reaches a 100\% overflow rate beyond 120 pages. Under the more common 128K deployment, the limitation becomes even more severe: 2048\,px starts to overflow at 40--60 pages, and 1568\,px overflows for all documents longer than 80 pages. In contrast, 512\,px never overflows under either setting.
\begin{table}[h]
\centering

\begin{tabular}{lrrrrr}
\toprule
\textbf{Length (pages)} & \textbf{$n$} & \textbf{512} & \textbf{1024} & \textbf{1568} & \textbf{2048} \\
\midrule
\multicolumn{6}{l}{\emph{256K context}}\\
0--80    & 954 & 0.0 & 0.0 & 0.0   & 0.0 \\
80--120  &  81 & 0.0 & 0.0 & 0.0   & \textbf{79.0} \\
120+     &  56 & 0.0 & 8.9 & 100.0 & 100.0 \\
\midrule
\multicolumn{6}{l}{\emph{128K context}}\\
0--40    & 665 & 0.0 & 0.0 & 0.0   & 0.0 \\
40--60   & 153 & 0.0 & 0.0 & 0.0   & 51.6 \\
60--80   & 136 & 0.0 & 0.0 & 52.9  & 90.4 \\
80--120  &  81 & 0.0 & 0.0 & 100.0 & 100.0 \\
120+     &  56 & 0.0 & \textbf{85.7} & 100.0 & 100.0 \\
\bottomrule
\end{tabular}
\caption{
Context-overflow rate (\%) on MMLongBench-Doc under different page resolutions
and context-window sizes.}
\label{tab:overflow}
\end{table}

These results directly motivate the hierarchical design of DocTrace. Stage~1 uses 512\,px page images to localize evidence over the entire document without context overflow, while Stage~3 revisits only the retrieved evidence pages at 1568\,px for fine-grained reasoning. This coarse-to-fine design enables high-resolution document understanding without sacrificing scalability.

\section{Stage-wise Error Analysis}

To better understand the remaining errors of DocTrace, we attribute prediction failures to different stages of the pipeline. We analyze answerable and unanswerable questions separately. For answerable questions, we study whether errors arise from incomplete evidence localization (Stage~1) or reasoning over retrieved evidence (Stage~3). For unanswerable questions, we analyze how hallucinations relate to the Stage~1 answerability decision and retrieved distractor pages.

Table~\ref{tab:err-answerable} shows that Stage~3 generalizes well once the required evidence has been retrieved. When all gold evidence pages are available, single-page and multi-page questions achieve nearly identical accuracy (61.7 vs.~60.2), indicating that reasoning itself is not the primary limitation. The overall performance gap (52.6 vs.~40.0) is instead explained by the much lower Stage~1 coverage on multi-page questions (46.5 vs.~78.3). Moreover, partial retrieval remains substantially better than retrieving no evidence (26.7 vs.~11.2), suggesting that Stage~3 effectively exploits whatever evidence is available. Overall, the performance degradation on multi-page questions is primarily caused by incomplete evidence localization.
\begin{table}[h]
\centering

\begin{tabular}{lcccccc}
\toprule
\textbf{Type}
&
$\emph n$
&
\textbf{Cov.}
&
\textbf{Full}
&
\textbf{None}
&
\textbf{Part.}
&
\textbf{Overall}
\\
\midrule

Single
&
475
&
78.3
&
\textbf{61.7}
&
20.0
&
---
&
52.6
\\

Multi
&
355
&
46.5
&
\textbf{60.2}
&
11.2
&
26.7
&
40.0
\\

\bottomrule
\end{tabular}

\caption{
Performance by retrieval completeness.
\textbf{Cov.} denotes Stage~1 coverage ($\mathrm{Recall}=1$).
\textbf{Full}, \textbf{None}, and \textbf{Part.} denote the answer accuracy when all, none, or part of the gold evidence pages are retrieved, respectively.
}

\label{tab:err-answerable}
\end{table}

Table~\ref{tab:err-unanswerable} attributes hallucinations to Stage~1. All
hallucinations originate from questions that Stage~1 incorrectly admits as
answerable. Whenever Stage~1 correctly rejects an unanswerable question,
hallucination never occurs.

Although Stage~3 recovers 66.8\% of the wrongly admitted cases by predicting a refusal, it cannot completely eliminate the errors. Furthermore, hallucination increases monotonically with the number of retrieved distractor pages (0.0\%, 28.8\%, 39.7\%, and 44.4\%), indicating that irrelevant retrieved pages make the downstream model increasingly likely to generate unsupported answers.

\begin{table}[h]
\centering
\begin{tabular}{lcc}
\toprule
\textbf{Setting} & $\emph n$ & \textbf{Halluc. (\%)}\\
\midrule
\multicolumn{3}{l}{\textit{Stage-1 decision}}\\
Rejected
&
24
&
\textbf{0.0}
\\
Accepted
&
220
&
33.2
\\
\quad Recovered (Stage~3 refused)
&
147
&
0.0
\\
\quad Not recovered (hallucinated)
&
73
&
100.0
\\
\midrule
\multicolumn{3}{l}{\textit{Retrieved pages}}\\
0
&
24
&
\textbf{0.0}
\\
1
&
139
&
28.8
\\
2
&
63
&
39.7
\\
$\ge3$
&
18
&
44.4
\\
\midrule
Overall
&
244
&
29.9
\\
\bottomrule
\end{tabular}
\caption{
Hallucination analysis on unanswerable questions. Within the Accepted subset, we further break down outcomes into cases where Stage~3 recovers by predicting a refusal versus cases that result in a hallucinated answer.
}
\label{tab:err-unanswerable}
\end{table}

Both analyses consistently identify Stage~1 as the primary bottleneck of
DocTrace. For answerable questions, the performance gap on multi-page reasoning
is mainly caused by incomplete evidence localization rather than reasoning
errors. For unanswerable questions, hallucinations originate from incorrect
Stage~1 answerability decisions and become more frequent as more distractor
pages are retrieved. These findings suggest that improving Stage~1 retrieval
recall and answerability prediction is likely to provide the largest overall
performance gain.

\section{Baseline Details}

\subsection{Evaluation Protocol}

This section provides additional details of the compared baselines and clarifies the evaluation protocol adopted in our experiments.

For \textbf{MMLongBench-Doc}, the performance of proprietary MLLMs, including GPT-4.1, GPT-4o, Claude-3.7-Sonnet, and Gemini-1.5-Pro, is directly taken from the official MMLongBench-Doc leaderboard. These models are evaluated under the unified protocol provided by the benchmark, ensuring fair comparison across different proprietary systems.

For the remaining baseline methods, including representative end-to-end MLLMs, retrieval-augmented approaches, and agent-based methods, we report the results published in their original papers whenever they are evaluated on the same benchmark.

For \textbf{DocTrace} and the \textbf{Qwen3-VL-8B-Instruct} backbone, all experiments are conducted by ourselves following the official evaluation scripts and protocols released by each benchmark. Specifically, we strictly follow the official evaluation procedures of MMLongBench-Doc, LongDocURL, and SlideVQA without introducing any task-specific modifications, ensuring reproducible and fair comparisons.

\subsection{End-to-End MLLMs}

End-to-end MLLMs directly process the complete document without performing explicit evidence retrieval or intermediate page selection. Evidence localization and multi-page reasoning are implicitly handled within the model's long-context representations, making these approaches heavily dependent on large context windows and strong multimodal reasoning capability.

We compare against representative end-to-end document understanding models, including mPLUG-DocOwl2, Docopilot, InternVL3, and DocSeeker.

\paragraph{mPLUG-DocOwl2}

mPLUG-DocOwl2 is an OCR-free document understanding model that directly aligns document images with a large language model through a dedicated visual abstraction module. By learning unified visual-text representations via large-scale instruction tuning, it performs document reasoning without relying on external OCR systems.

\paragraph{Docopilot}

Docopilot follows a retrieval-free paradigm that directly processes complete document images within a single multimodal model. It combines efficient long-context attention mechanisms with multimodal data packing, enabling high-resolution document understanding while avoiding explicit evidence retrieval.

\paragraph{InternVL3}

InternVL3 is a general-purpose multimodal large language model featuring native multimodal pre-training and strong OCR capability. Efficient visual position encoding together with advanced post-training strategies enables competitive long-context document understanding performance across diverse multimodal benchmarks.

\paragraph{DocSeeker}

DocSeeker emphasizes structured visual reasoning and evidence grounding for long document understanding. It exploits layout-aware document representations to improve evidence localization and cross-page reasoning while maintaining an end-to-end inference framework.

\subsection{Retrieval-Augmented Methods}

Retrieval-augmented methods first retrieve a subset of relevant document pages or visual regions before performing answer generation. By restricting expensive multimodal reasoning to retrieved evidence only, these approaches substantially improve inference efficiency compared with processing the complete document.

Our comparison includes representative visual retrieval methods, namely M3DocRAG, VisRAG, SV-RAG, VDocRAG, MoLoRAG, and URaG.

\paragraph{M3DocRAG}

M3DocRAG formulates long document understanding as a retrieval-augmented generation problem. It retrieves relevant document pages through multimodal retrieval and performs answer generation only on the selected evidence, reducing the reasoning space for long documents.

\paragraph{VisRAG}

VisRAG performs retrieval directly in the visual domain by representing document pages as images rather than OCR text. It employs a dual-encoder retriever to identify relevant pages, followed by a vision-language model that generates answers from the retrieved visual evidence.

\paragraph{SV-RAG}

SV-RAG unifies retrieval and answer generation within a single multimodal backbone using two specialized LoRA adapters. One adapter is optimized for evidence retrieval through contrastive learning, while the other performs autoregressive answer generation.

\paragraph{VDocRAG}

VDocRAG is designed for visually rich documents by learning dense visual representations for page retrieval. Retrieved page images are then passed to a multimodal generator for answer prediction, avoiding explicit conversion of document pages into textual representations.

\paragraph{MoLoRAG}

MoLoRAG introduces logic-aware multimodal retrieval that explicitly considers reasoning dependencies during evidence selection. By jointly modeling retrieval and logical relevance, it improves evidence quality for complex multi-page reasoning.

\paragraph{URaG}

URaG proposes a unified retrieval-generation framework that jointly optimizes evidence retrieval and answer generation within a single training objective, enabling more effective interaction between retrieval and reasoning.

\subsection{Agent-based Methods}

Agent-based methods formulate long document understanding as an interactive decision-making process. Instead of processing all document pages simultaneously, the model iteratively explores the document through search, navigation, or perception actions, progressively collecting evidence before generating the final answer.

We compare against representative document agents, including VRAG-RL, Doc-V*, and MM-Doc-R1.

\paragraph{VRAG-RL}

VRAG-RL formulates long document reasoning as a sequential decision-making problem. During inference, the agent alternates between reasoning and visual perception actions to progressively collect evidence from the document. The policy is optimized using GRPO with rewards encouraging both accurate retrieval and correct answer prediction.

\paragraph{Doc-V*}

Doc-V* adopts a coarse-to-fine interactive reasoning strategy for multi-page document understanding. The agent progressively narrows the search space through iterative exploration and evidence inspection before producing the final answer.

\paragraph{MM-Doc-R1}

MM-Doc-R1 trains document agents through reinforcement learning for long document visual question answering. Rather than relying solely on supervised instruction tuning, it optimizes multi-turn interaction policies that iteratively retrieve and aggregate evidence before answer generation.

Although these agent-based approaches expose intermediate interaction trajectories, their reasoning processes remain action-oriented. In contrast, DocTrace explicitly organizes grounded document evidence into executable evidence graphs, allowing every intermediate reasoning step to be directly traced back to supporting document evidence and providing explicit node-level provenance.

\end{document}